\documentclass[letterpaper, 10 pt, conference]{ieeeconf}  

\IEEEoverridecommandlockouts                              

\usepackage{times}
\usepackage{epsfig}
\usepackage{graphicx}
\usepackage{amsmath}
\usepackage{amssymb}
\usepackage{adjustbox}
\usepackage{bm}
\usepackage{multirow}
\usepackage{booktabs}
\usepackage{color}
\usepackage{cite}
\title{\LARGE \bf
	Cascaded Non-local Neural Network for Point Cloud Semantic Segmentation
}

\author{
	Mingmei Cheng, Le Hui, Jin Xie$^*$, Jian Yang and Hui Kong
	\thanks{
		The authors are with PCA Lab, Key Lab of Intelligent Perception and Systems for High-Dimensional Information of Ministry of Education, and Jiangsu Key Lab of Image and Video Understanding for Social Security, School of Computer Science and Engineering, Nanjing University of Science and Technology, Nanjing, China. 210094. Email: \{chengmm, le.hui, csjxie, konghui, csjyang\}@njust.edu.cn
	}%
	\thanks{*The author responsible for the correspondence of this paper}%
	\thanks{This work was supported by the National Science Fund of China (Grant Nos. U1713208, 61876084), Program for Changjiang Scholars.}%
}

\begin{document}

\maketitle
\thispagestyle{empty}
\pagestyle{empty}

\begin{abstract}

	In this paper, we propose a cascaded non-local neural network for point cloud segmentation. The proposed network aims to build the long-range dependencies of point clouds for the accurate segmentation. Specifically,  we develop a novel cascaded non-local module, which consists of the neighborhood-level, superpoint-level and global-level non-local blocks. First, in the neighborhood-level block, we extract the local features of the centroid points of point clouds by assigning different weights to the neighboring points. The extracted local features of the centroid points are then used to encode the superpoint-level block with the non-local operation. Finally, the global-level block aggregates the non-local features of the superpoints for semantic segmentation in an encoder-decoder framework. Benefiting from the cascaded structure, geometric structure information of different neighborhoods with the same label can be propagated. In addition, the cascaded structure can largely reduce the computational cost of the original non-local operation on point clouds. Experiments on different indoor and outdoor datasets show that our method achieves state-of-the-art performance and effectively reduces the time consumption and memory occupation.

\end{abstract}

\section{INTRODUCTION}

	In recent years, 3D imaging sensors have been greatly developed to facilitate the acquisition of 3D point cloud data. With the explosive growth of 3D point cloud data, point cloud semantic segmentation has received more and more attention~\cite{pc_ye20183d, pc_engelmann2017exploring, intro_li2018pointcnn} in 3D scene understanding. Point cloud semantic segmentation aims to classify each point into a category. Due to the unordered and irregular structure of 3D point clouds, how to exploit context information of point clouds for semantic segmentation is very challenging.

	Recently,  various efforts have been made on point cloud semantic segmentation. PointNet~\cite{pc_qi2017pointnet} directly employs the multi-layer perception (MLP) to extract the local feature of a single point for point cloud segmentation. Based on PointNet, PointNet++~\cite{qi2017pointnet++} and DGCNN~\cite{pc_wang2019dynamic} aggregate  different local features of point clouds with the max pooling operation for segmentation. Based on the PointNet++ framework, PointWeb~\cite{pc_zhao2019pointweb} learns the weights between each pair of points in a local region to extract the local features. Although these methods can capture the geometric structures of different neighborhoods well, the relationships between long-range neighborhoods of point clouds are ignored.

	In fact, the context from long-range neighboring points is important for point cloud semantic segmentation.  It is desirable to exploit the long-range dependencies of different neighborhoods to characterize  discriminative geometric structures of point clouds.
	To this end, we propose a novel cascaded non-local neural network for segmentation. In our method, the non-local operation is performed on three levels, i.e., the neighborhood level, superpoint level and global level, corresponding to different scales of areas in point clouds. The non-local operation at the neighborhood level is applied to the neighboring points with the $k$-nearest neighbors ($k$-NN) algorithm.  The non-local operation at the superpoint-level is conducted in the superpoint area, which is  a set of points with the isotropous geometric characteristics, while the  global-level operation is implemented in the global point clouds composed of all the superpoints. By cascading the three non-local operations, geometric structure information of neighboring points can be propagated level by level. Once the non-local features at the global level are obtained, we employ the encoder-decoder framework in PointNet++ to predict the label of each point for semantic segmentation. Experimental results on the S3DIS, ScanNet, vKITTI and SemanticKITTI datasets demonstrate the effectiveness of our proposed method.

	In the cascaded non-local neural network, the non-local operation, as an attention mechanism, can learn the attention weights of neighboring points at each level. Thus, different weights can be highlighted for different contexts from long-range neighboring points. In addition, the original non-local operation performed on the whole point clouds is time-consuming. By partitioning the whole point clouds into the cascaded areas, the resulting cascaded non-local operation can largely reduce the computational cost of the original non-local module.

	In summary, the main contributions of this paper are two-fold. On one hand, we develop a novel cascaded non-local module where the non-local features of the centroid points at different levels can be extracted. Thus, context information from long-range neighboring points can be propagated level by level. On the other hand, the cascaded non-local module can largely reduce the computational complexity of the original non-local operation.

\section{Related Work}
\subsection{Deep learning on point clouds}

PointNet~\cite{pc_qi2017pointnet} is the pioneering algorithm to extract deep features of unordered point clouds. PointNet++ \cite{qi2017pointnet++} pays more attention to extract the local feature of point clouds with the hierarchical structure. PointWeb \cite{pc_zhao2019pointweb} learns the weights of pairwise points to enhance the local features of point clouds. LatticeNet \cite{ rosu2019latticenet} adopts a sparse permutohedral lattice to to characterize the local features of point clouds.

Graph-based methods mainly focus on depicting edge relationships between points, and thus boost the local feature embedding. DGCNN \cite{pc_wang2019dynamic} constructs the $k$-nn graph to characterize the local geometric structures of point clouds so that the local features of the points can be extracted. GACNet~\cite{gac} introduces a graph attention convolution to assign weights to neighboring points and extracts the local features of point clouds through an attention mechanism. Models based on the superpoint graph (SPG) framework~\cite{landrieu2018large, re_landrieu2019point} partition point clouds into superpoints, and then conduct feature embedding through a graph neural network built upon SPG. 

SqueezeSegV3 \cite{xu2020squeezesegv3} adopts spherical projection to generate LiDAR images and proposes the spatially-adaptive convolution for point cloud feature embedding, where the spatial priors in the LiDAR images can be exploited. SalsaNet \cite{aksoy2019salsanet} is an encoder-decoder network for point cloud segmentation, where the bird-eye view image is used as the input and an auto-labeling process is employed to transfer the labels from the camera to the LiDAR. Due to the imbalanced spatial distribution PolarNet \cite{zhang2020polarnet} proposes the polar bird’s-eye-view representation so that the nearest-neighbor-free segmentation method is used for point cloud semantic segmentation.

\subsection{Non-local neural networks}

The idea of the non-local operation was originally introduced in image denoising \cite{buades2005non}. Wang {\em et~al.} \cite{re_wang2018non} propose a neural network combining the non-local blocks with CNNs to extract the image features. However, the vast computational consumption and massive memory occupation hinder its application. In \cite{re_zhu2019asymmetric}, Zhu {\emph{et al.}} propose an asymmetric non-local neural network to reduce the computational cost through a pyramid sampling module. Zhang {\em et~al.} \cite{re_zhang2019latentgnn} introduce a novel graph structure to reduce the computational complexity of the non-local operation. In \cite{nonlocal_huang2019ccnet}, only the pixels along the criss-cross path participate in the non-local operation. In \cite{nonlocal_kraning2014dynamic}, an adaptive sampling strategy is adopted to decrease the computational complexity. These methods aiming at reducing the computational cost of the non-local operation are based on the regular structure of images, which cannot be directly extended to the irregular and unordered point clouds for point cloud feature extraction.

\section{Our Method}
In this section, we present our cascaded non-local neural network for point cloud segmentation. In Sec.~\ref{sec:revisiting_the_nonlocal}, we briefly revisit the non-local operation. In Sec.~\ref{sec:cascaded_nonlocal_module}, we describe the details of our cascaded non-local module. Finally, we present our network architecture in Sec. \ref{sec:architecture} and analyze the computational complexity of our cascaded non-local module in Sec.~\ref{sec:complexity}.

\subsection{Revisiting the non-local operation}\label{sec:revisiting_the_nonlocal}
Recently, the non-local operation is employed to construct the deep non-local  network~\cite{re_wang2018non}. Given the feature map $\mathbf{x}\in \mathcal{R}^{H \times W \times C}$ of an image, we denote $H$, $W$, and $C$ as the height and width of the image and the number of the channels, respectively. Suppose that in each non-local block there are three convolutions $\bm{\theta}=W_{\theta}(\mathbf{x})$, $\phi\bm{\phi}=W_{\phi}(\mathbf{x})$, and $\bm{\gamma}=W_{\gamma}(\mathbf{x})$ for embedding, respectively, where $\bm{\theta}\in\mathcal{R}^{H\times W\times C'}$, $\bm{\phi}\in\mathcal{R}^{H\times W\times C'}$, $\bm{\gamma}\in\mathcal{R}^{H\times W\times C'}$ and $C'$ is the number of the output channels. We then reshape these three embeddings to $\bm{\theta}\in\mathcal{R}^{N\times C'}$, $\bm{\phi}\in\mathcal{R}^{N\times C'}$ and $\bm{\gamma}\in\mathcal{R}^{N\times C'}$, where $N=H\times W$. The final non-local feature is calculated as:
\begin{equation}
\setlength{\abovedisplayskip}{0pt}
\setlength{\belowdisplayskip}{2pt}
\mathbf{y}=softmax(\bm{\theta}\times\bm{\phi}^{\top})\times\bm{\gamma} \label{eq_orig_nonlocal}
\end{equation}
where $\mathbf{y} \in \mathcal{R}^{N\times C'}$ is the weighted sum of the embeddings of all pixels.

However, the non-local operation leads to the high computational cost due to large amounts of multiplications in Eq. \ref{eq_orig_nonlocal}. To tackle this problem, existing methods mainly focus on sampling the feature maps to reduce the computational complexity of matrix multiplications such as~\cite{nonlocal_huang2019ccnet, nonlocal_kraning2014dynamic, re_zhu2019asymmetric}. Due to the regular grid structure of images, these methods can effectively reduce the computational cost of the non-local operation. Nonetheless, since point clouds are irregular and unordered, it is difficult to regularly sample the feature maps of point clouds. Therefore, these methods cannot be directly applied to extract non-local features of point clouds.

\setlength{\abovecaptionskip}{0.cm}
\begin{figure}[t]
	\begin{center}
		\includegraphics[width=0.95\linewidth]{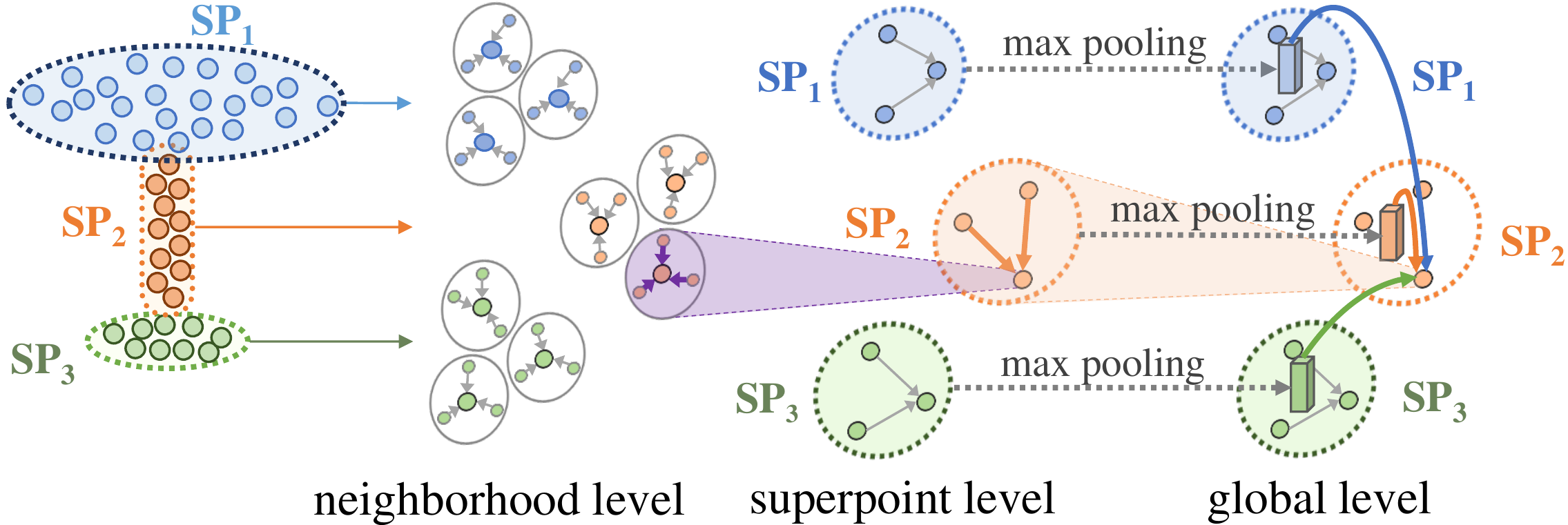}
	\end{center}
	\vspace{-4pt}
	\caption{ The non-local operation at three levels in our cascaded nonlocal module, including the neighborhood level, superpoint level and global level.
	}
	\label{fig:non-local_area}
\end{figure}
\setlength{\belowcaptionskip}{-0.cm}

\setlength{\abovecaptionskip}{0.cm}
\begin{figure*}[t]
	\begin{center}
		\includegraphics[width=0.85\linewidth]{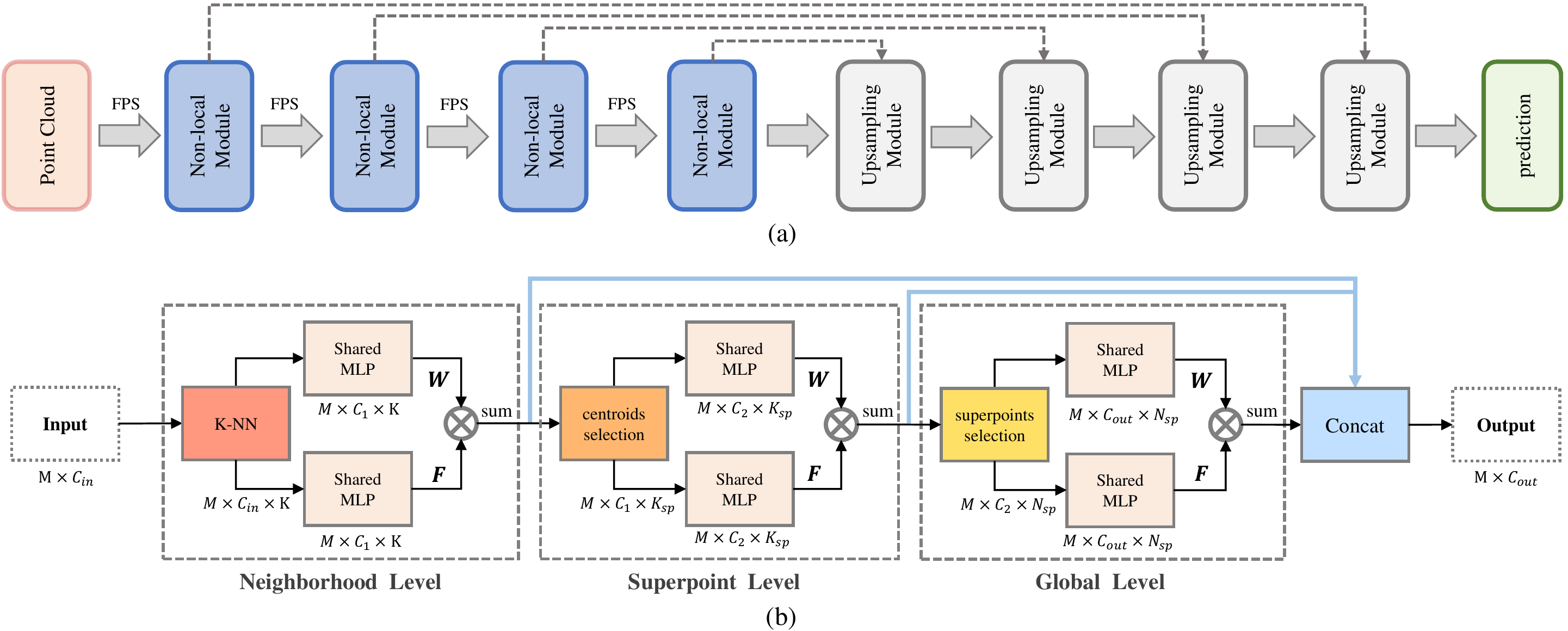}
	\end{center}
	\vspace{-6pt}
	\caption{
		(a) Framework of our cascaded non-local neural network. Four non-local modules are employed to construct the encoder network for point clouds feature embedding. The decoder is composed of four upsampling modules to predict the semantic labels of points. (b) Illustration of the proposed non-local module. The neighborhood-level non-local operation is performed to extract the local features of the neighborhood constructed by $K$ nearest neighboring points.
		Then the superpoint-level non-local operation is conducted to propagate the geometric features of various neighborhoods to the superpoint. 
		Consequently, the global-level non-local operation is performed on various superpoints, which further propagates the superpoint level non-local features in the global area. Finally, the non-local features at three levels are concatenated to generate discriminative features for point cloud semantic segmentation.}
	\label{fig:framework}
\end{figure*}
\setlength{\belowcaptionskip}{-0.cm}

\subsection{Cascaded non-local module}\label{sec:cascaded_nonlocal_module}

Given two pointwise features ${\bm f}_{i}\in \mathbb{R}^C$ and ${\bm f}_{j}\in \mathbb{R}^C$, the non-local operation for point clouds is formulated as follows:
\begin{equation}
\setlength{\abovedisplayskip}{6pt}
\setlength{\belowdisplayskip}{6pt}
{\bm o}_{i}=\sum \nolimits _j h({\bm f}_{i}, {\bm f}_{j})* g({\bm f}_{j})
\label{equ:equ_nonlocal}
\end{equation}
where $h({\bm f}_{i}, {\bm f}_{j}) = \bm{W}^{\top}_{\theta}({\bm f}_{i}-{\bm f}_{j})$ is the pairwise function embedding the difference between the two feature vectors, $g({\bm f}_{j})=\bm{W}^{\top}_{\phi}({\bm f}_j)$ is a unary function for feature embedding, the operator $*$ represents the Hadamard product. $\bm{W}_{\theta}\in\mathbb{R}^{C\times D}$ and $\bm{W}_{\phi}\in\mathbb{R}^{C\times D}$ are the two weights to be learned.


In Eq. \ref{eq_orig_nonlocal}, the original non-local function employs the scalar value to depict the similarity between each pair of points. However, our pairwise function employs a channel-wise vector $h({\bm f}_{i}, {\bm f}_{j})\in\mathbb{R}^{1\times D}$ to describe the relationship between two points. Thus our non-local feature can capture the geometry structure more accurately. Furthermore, to deal with the feature scale of different pair of points, the pairwise function $h$ is normalized as follow:
\begin{equation}
\setlength{\abovedisplayskip}{6pt}
\setlength{\belowdisplayskip}{6pt}
h^{d}({\bm f}_{i}, {\bm f}_{j})=\frac{exp(h^{d}({\bm f}_{i}, {\bm f}_{j}))}{\sum_qexp(h^{d}({\bm f}_{i}, {\bm f}_{q}))}
\end{equation}
where $d=1,2,\cdots,D$ represents the $d$-th channel of the feature vector.

It is noted that it is infeasible to directly perform the non-local operation on the whole point clouds since the non-local operation defined in Eq.~\ref{equ:equ_nonlocal} requires the huge memory occupancy. Therefore, in order to balance the memory cost of the non-local operation and accurate depiction of geometric structures of point clouds, we propose a cascaded non-local module, where the nonlocal operation can be performed on the point clouds at different levels. Specifically, the three-level non-local operations are conducted in three different scales of areas: neighborhood area, superpoint area and global area, respectively. It can greatly reduce the computational complexity of the non-local opeation by controlling the number of points participated in each non-local operation.

\textbf{Neighborhood level} The non-local operation at the neighborhood level aims to extract the local features of the centroids in the corresponding neighborhoods. We first leverage the farthest point sampling (FPS) to choose $M$ points as the centroids. For each centroid $\bm{p}_i$, the $K$ nearest neighboring points are used to construct the neighborhood area (Fig. \ref{fig:non-local_area}). Consequently, we apply the non-local operation in the neighborhood area to extract the local feature of the centroid:
\begin{equation}
\setlength{\abovedisplayskip}{5pt}
\setlength{\belowdisplayskip}{5pt}
{\bm x}_{i}=\sum \nolimits _{j=1}^{K} h({\bm f}_{i}, {\bm f}_{j})*g({\bm f}_{j})
\label{eq:eq_nonlocal1}
\end{equation}
where $i = 1,2,\cdots,M$ is the index of the centroid points. From Eq.~\ref{eq:eq_nonlocal1}, one can see that the local feature of each centroid point can be characterized by assigning different weights to the points in the neighborhood.

\textbf{Superpoint level} Once the local features of the centroid points are extracted, we conduct the non-local operation at the superpoint level. It is expected that geometric structure information of different neighborhoods in the superpoint can be effectively propagated. Superpoint is a set of points with isotropically geometric features. Generally, the number of centroid points in the superpoint is different. Therefore, in order to facilitate batch processing in our neural network, we randomly sample $K_{sp}=20$ centroid points in the superpoint. Thus, for the centroid point $\bm{p}_i$, the non-local feature ${\bm y}_{i}$ at the superpoint level is defined as:
\begin{equation}
\setlength{\abovedisplayskip}{7pt}
\setlength{\belowdisplayskip}{7pt}
{\bm y}_{i}=\sum \nolimits _{j=1}^{K_{sp}} h({\bm x}_{i}, {\bm x}_{j})*g({\bm x}_{j})
\end{equation}
where ${\bm x}_{i}$ is the local feature of the centroid point $\bm{p}_i$ at the neighborhood level.


\textbf{Global level} In order to exploit semantic contexts from different superpoints in the point clouds, we furthermore propagate the features of the centroid points at the superpoint level. Since each superpoint contains multiple centroids, we use a max pooling operation to extract the superpoint feature ${\bm v}_{j}$ of the $j$-th superpoint. As shown in Fig.~\ref{fig:non-local_area}, each small box represents the corresponding superpoint feature. For the centroid point $\bm{p}_i$, by assigning different weights to the superpoint features, we define the following non-local operation at the global level:
\begin{equation}
\setlength{\abovedisplayskip}{7pt}
\setlength{\belowdisplayskip}{7pt}
{\bm z}_{i}=\sum \nolimits ^{N_{sp}}_{j=1} h({\bm y}_{i}, {\bm v}_{j})*g({\bm v}_{j})
\end{equation}
where $N_{sp}$ is the number of superpoints. 

After propagating the non-local features of the centroids through the whole point clouds, we can obtain the fused features with a mapping function $M_{\gamma} :\mathbb{R}^D \rightarrow  \mathbb{R}^{D^{+}}$. Therefore, the final feature of each centroid point is formulated as:
\begin{equation}
\setlength{\abovedisplayskip}{6pt}
\setlength{\belowdisplayskip}{6pt}
\mathcal{X}_i=M_{\gamma}([{\bm x}_{i},\; {\bm y}_{i},\; {\bm z}_{i}])
\end{equation}
where $[\cdot,\cdot,\cdot]$ denotes the concatenation operation and ${\bm x}_{i}$, ${\bm y}_{i}$, ${\bm z}_{i}$ represent the non-local features at three levels, respectively. With the cascaded non-local operations at three levels, the long-range dependencies across different neighboring points can be build so that we can obtain the discriminative feature of each centroid point for semantic segmentation.

\subsection{Architecture}\label{sec:architecture}
The overall architecture of our network is illustrated in Fig.~\ref{fig:framework}. Our network is constructed based on the PointNet++ framework, combing the cascaded non-local modules to build the long-range dependencies between the points. In the encoder, we employ four non-local modules for feature embedding while in the decoder we adopt the upsampling strategy in PointNet++ to predict the semantic labels of the points.

In PointNet++, the local features of the centroid points are extracted by performing the max pooling operation on the features of single points in a local region. Different from PointNet++, our proposed method extracts more discriminative local features by weighting the neighboring points through the non-local mechanism. In addition, we propagate the local features of the centroid points at the superpoint and global levels with the non-local operation. Thus, the discriminativeness of the local features of the centroid points can be further boosted. On the contrary, PointNet++ mainly focuses on the hierarchical local regions for feature extraction without considering non-local regions such as superpoints.

\subsection{Computational complexity analysis}\label{sec:complexity}

In Eq.~\ref{equ:equ_nonlocal}, the non-local feature of each point is computed with the weighted sum of responses of all $N$ points in the point clouds. If we ignore the feature channel, the computational complexity is $O(N^2)$. For our cascaded non-local module, the non-local operation is performed at three levels and the number of points participated in each operation is far smaller than $N$. Specifically, in the neighborhood level, the number of the subsampled centroid points is $M$ ($M=1024,256,64,16$ in the four non-local modules) and each centroid point has $K$ neighboring points. At the superpoint level, for each centroid point, the non-local operation is performed on $K_{sp}$ centroid points in the same superpoint. At the global level, $N_{sp}$ superpoints are involved in the non-local operation for each centroid point. In the experiment, we set $N_{sp} \leqslant 32$. Therefore the final computational complexity $T$ of our hierarchical nonlocal operation is:
\begin{equation}
\setlength{\abovedisplayskip}{7pt}
\setlength{\belowdisplayskip}{7pt}
T=O(M\cdot K) + O(M\cdot K_{sp}) + O(M\cdot N_{sp})
\end{equation}
where $K$, $K_{sp}$ and $N_{sp}$ are far smaller than $N$ and $M < N$. Thus, we can significantly reduce the computational complexity of the non-local operation on point clouds.

\setlength{\abovecaptionskip}{0.cm}
\begin{table}[]
	\begin{center}
		\begin{adjustbox}{width=0.95\linewidth}
			\begin{tabular}{l|ccc|c|ccc}
				\midrule
				\multirow{2}{*}{Method} & &S3DIS & &ScanNet &\multicolumn{3}{c}{vKITTI}\\
				&mIoU&mAcc&OA &OA &mIoU&mAcc&OA \\
				\midrule
				PointNet~\small\cite{pc_qi2017pointnet}	&47.6&66.2&78.5  &73.9  &34.4&47.0&79.7\\
				PointNet++~\small\cite{qi2017pointnet++}&54.5&67.1&81.0  &84.5  &-&-&-\\
				SPGragh~\small\cite{landrieu2018large} &62.1&73.3&85.5 &- &-&-&-\\				
				ShellNet~\small\cite{zhang2019shellnet} &66.8 &- &87.1    &85.2    &-&-&-\\
				G+RCU~\small\cite{pc_engelmann2017exploring} &49.7  &66.4  &81.1    &-      &35.6&57.6&79.7\\
				G+RCU2~\small\cite{engelmann2018know}&58.3  &67.8  &84.0      &-    &36.2&49.7&80.6\\
				PointCNN~\small\cite{intro_li2018pointcnn} &65.4 &75.6 &88.1    &85.1     &-&-&-\\
				PointWeb~\small\cite{pc_zhao2019pointweb} &66.7 &76.2 &87.3     &85.9    &-&-&-\\			
				SSP+SPG~\small\cite{re_landrieu2019point}&{\bf68.4} &{\bf78.3} &87.9    &-     &52.0&67.3&84.3\\
				HEPIN~\small\cite{pointedge2019}&67.8&76.3&88.2 &- &-&-&- \\
				ConvPoint~\small\cite{boulch2019generalizing}&64.7&-&87.9 &- &-&-&-\\
				3P-RNN~\small\cite{pc_ye20183d}&56.2&73.4&86.9    &-     &41.6&54.1&87.8 \\
				\midrule
				PointNL (ours) &{\bf68.4}&77.5&{\bf88.2}&{\bf86.7} &{\bf 57.6} &{\bf 68.1}&{\bf 89.8}\\
				\midrule
			\end{tabular}
		\end{adjustbox}
	\end{center}
	\vspace{-2pt}
	\caption{Semantic segmentation results on S3DIS, ScanNet and vKITTI datasets.}
	\label{tab:results_various}
\end{table}
\setlength{\belowcaptionskip}{-0.cm}

\setlength{\abovecaptionskip}{0.cm}
\begin{figure}[t]
	\begin{center}
		\includegraphics[width=0.85\linewidth]{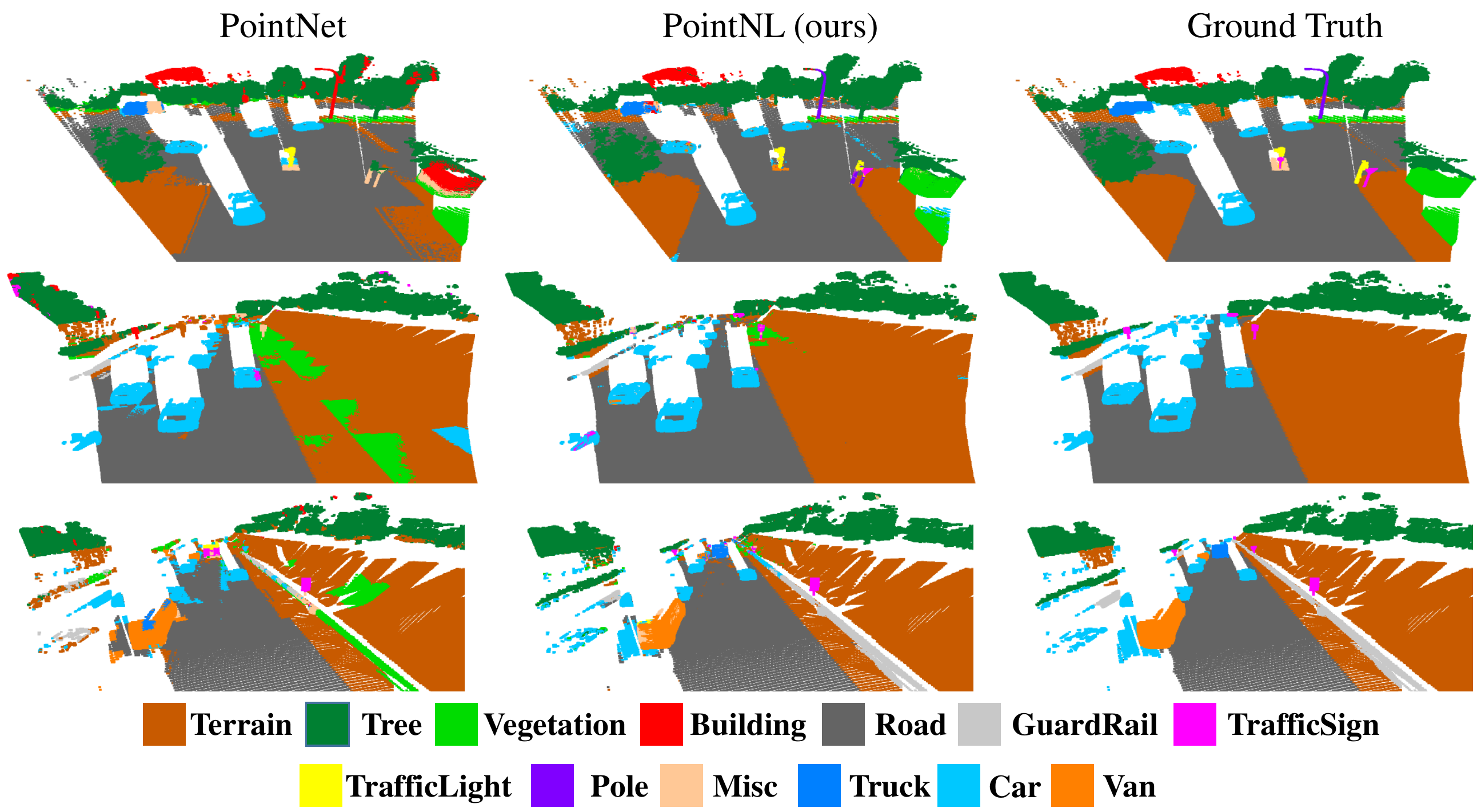}
	\end{center}
	\vspace{-1pt}
	\caption{Examples of semantic segmentation results on the vKITTI dataset.}
	\label{fig:vkitti}
\end{figure}
\setlength{\belowcaptionskip}{-0.cm}
\section{Experiments}
In this section, we evaluate our proposed model on indoor and outdoor datasets.

\setlength{\abovecaptionskip}{0.cm}
\begin{figure*}[t]
	\begin{center}
		\includegraphics[width=0.85\linewidth]{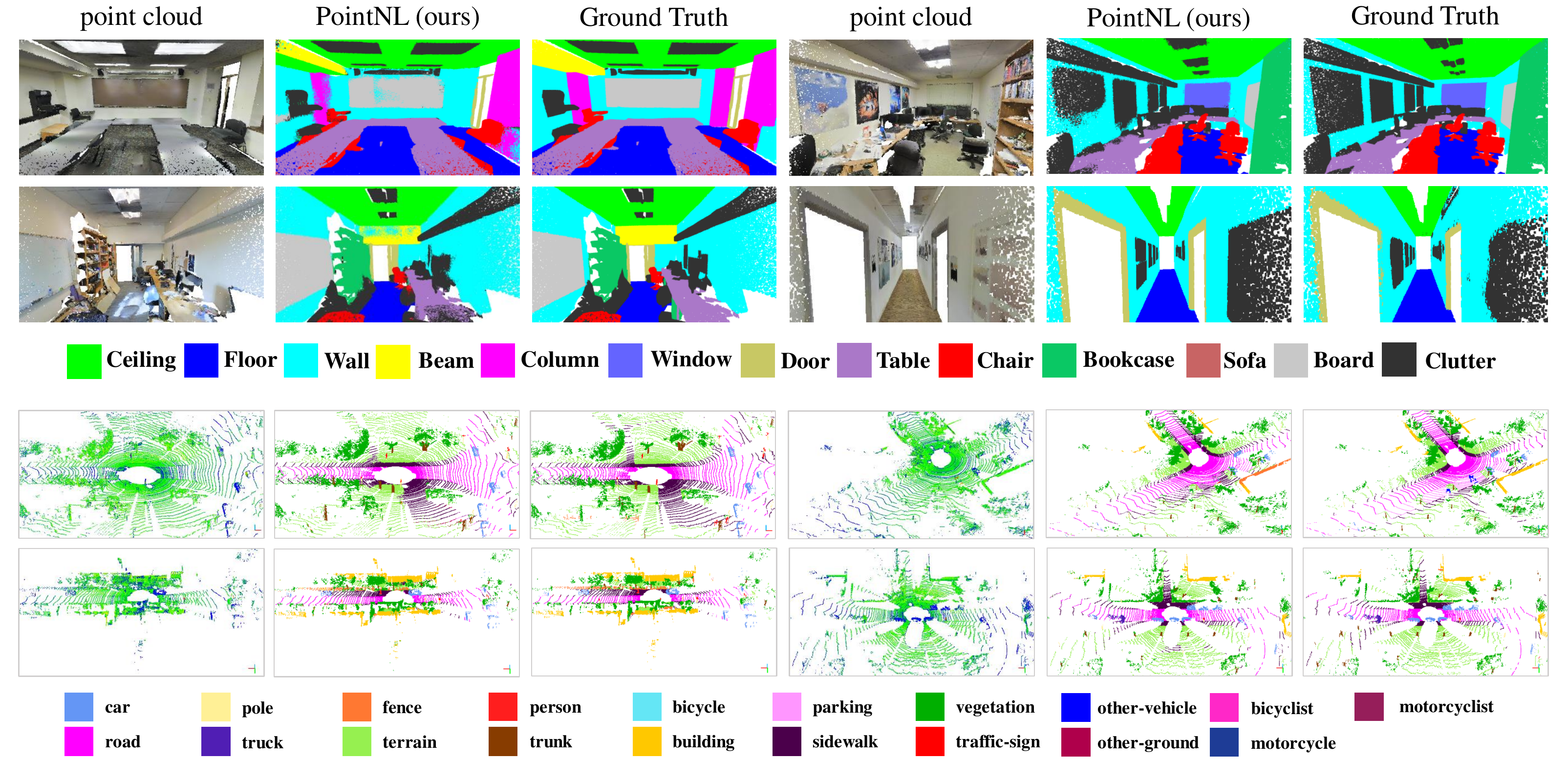}
	\end{center}
	\caption{Visualization results on the S3DIS and SemanticKITTI dataset, respectively. The results of the S3DIS are exhibited in the top two rows and the bottom two rows show the results of the SemanticKITTI.
	}
	\label{fig:results_s3dis_skitti}
\end{figure*}
\setlength{\belowcaptionskip}{-0.cm}

\subsection{Implementation details}
To train our model, we use the SGD optimizer, where the base learning rate and mini-batch size are set to 0.05 and 16, respectively. The momentum is set to 0.9 and the weight decay is 0.0001. The approach adopted for superpoints partition follows~\cite{landrieu2018large}. For the S3DIS~\cite{dataset_s3dis} dataset, the learning rate decays by 0.1 for every 25 epochs to train 100 epochs. For the ScanNet~\cite{dataset_scannet} dataset, we train the network for 200 epochs and decay the learning rate by 0.1 every 50 epochs. For the vKITTI~\cite{dataset_vkitti} dataset, we train the network for 100 epochs and decay the learning rate by 0.1 every 25 epochs. For the outdoor dataset SemanticKITTI~\cite{behley2019semantickitti}, the learning rate is 0.003 and weight decay is 0.0001. The scenes are partitioned into blocks as superpoints. Note that our method does not need to build the superpoint graphs.

\subsection{Semantic segmentation on datasets}
\textbf{S3DIS.} S3DIS~\cite{dataset_s3dis} is an indoor 3D point cloud dataset. The point clouds are split into six large-scale areas, where each point is annotated with one of the semantic labels from 13 categories. Three metrics are used to quantitatively evaluate our method: mean of per-class intersection over union (mIoU), mean of per-class accuracy (mAcc), and overall accuracy (OA). For evaluation, we follow the methods~\cite{qi2017pointnet++,landrieu2018large} to test the model on Area 5 and 6-fold cross validation. For training, we follow~\cite{qi2017pointnet++} to uniformly split the point clouds to blocks with an area size of 1m$\times$1m and randomly sample 4096 points in each block. During the test, we adopt all the points for evaluation. 

The quantitive results with 6-folds cross validation on S3DIS are shown in Tab. \ref{tab:results_various}. Our model named PoinNL can achieve better or comparable performance than other methods listed in Tab. \ref{tab:results_various}, benefiting from the rich features provided by the non-local module. It is noted that although additional hand-craft geometry features in \cite{landrieu2018large, re_landrieu2019point} are not utilized in our model, our PointNL can still achieve comparable results. The visual results are shown in Fig.~\ref{fig:results_s3dis_skitti}. It can be seen that our proposed method can obtain more accurate segmentation results on the S3DIS dataset.

\textbf{ScanNet.}
ScanNet~\cite{dataset_scannet} is an indoor scene dataset, which contains 1513 point clouds and points are annotated with 20 categories. Following \cite{qi2017pointnet++, pc_zhao2019pointweb}, the dataset is split into 1201 scenes for training and 312 scenes for testing. During the training, the points are divided into blocks of size 1.5m$\times$1.5m, and each block consists of 8192 points sampled on-the-fly. For testing, all the points in the test set are used for the evaluation. Following~\cite{pc_zhao2019pointweb}, the stride between the adjacent blocks is set as 0.5. 
The overall semantic voxel labeling accuracy (OA) is used to evaluate the compared segmentation methods.

We list the quantitative results on the validate dataset in Tab.~\ref{tab:results_various}. From this table, one can see that our model outperforms the other compared methods. Due to the hierarchical non-local module, which can capture the long-range context information of point clouds, our model can obtain good segmentation results.

\textbf{vKITTI.}
We also evaluate our method on the vKITTI~\cite{dataset_vkitti} dataset, which mimics the real-world KITTI dataset. There are 5 sequences of synthetic outdoor scenes and 13 classes (including road, tree, terrain, car, etc.) are annotated. The 3D point clouds are obtained by projecting the 2D depth image into 3D coordinates. For evaluation, we follow the strategy adopted in~\cite{pc_ye20183d} and split the dataset into 6 non-overlapping sub-sequences and conduct 6-fold cross validation. During the training, we split the point cloud into 4m $\times$ 4m blocks and randomly sample 4096 points in each block. For evaluation, the metrics, mean of per-class intersection over union (mIoU), mean of per-class accuracy (mAcc) and overall accuracy (OA) are adopted.

The quantitative results are listed in Tab.~\ref{tab:results_various}. From this table, one can see that our proposed PointNL can yield better performance on this dataset with a significant gain of $5.6\%$ in terms of mIoU. The visual results are shown in Fig. \ref{fig:vkitti}, which further demonstrates the effectiveness of our method.

\textbf{SemanticKITTI.}
SemanticKITTI~\cite{behley2019semantickitti} is a large-scale outdoor 3D dataset with 43552 LIDAR scans. Each scan contains $\sim 10^5$ points and can cover $160\times 160 \times 20$ meters in the 3D space. The dataset is split into 21 sequences, while the sequences 00$\sim$07 and 09$\sim$10 (19130 scans) are used for training, sequence 08 (4071 scans) for validation, and the sequences 11$\sim$21
(20351 scans) for online testing. The point clouds are annotated with 19 classes without color information. For evaluation, the mean of per-class intersection over union (mIoU) is adopted following~\cite{behley2019semantickitti}.

On SemanticKITTI dataset, PointNet~\cite{pc_qi2017pointnet}, PointNet++~\cite{qi2017pointnet++}, DarkNet21Seg~\cite{behley2019semantickitti}, DarkNet53Seg~\cite{behley2019semantickitti} can obtain the mIoUs of 14.6\%, 20.1\%, 47.4\% and 49.9\%, respectively, while our PointNL achieves the mIoU of 52.2\%.
It can be seen that our model outperforms the others. In addition, Fig.~\ref{fig:results_s3dis_skitti} visualizes the segmentation results on the SemanticKITTI dataset with our proposed PointNL.

\setlength{\abovecaptionskip}{0.cm}
\begin{table}	
	\small
	\begin{center}
		\begin{adjustbox}{width=0.85\linewidth}
			\begin{tabular}{l|ccc|cc}
				\toprule 
				Method&mIoU&mAcc&OA&Training time&GPU memory\\
				\midrule
				Non-local (baseline) &54.76&61.70&84.31 &6 days&60 G\\
				PointNL$^{1}$ (ours)&60.03&66.35&86.56 &1.4 days &5 G \\
				PointNL$^{12}$ (ours)&62.18&68.45&87.84 &1.6 days & 7 G \\			
				PointNL$^{123}$ (ours)&{\bf 63.50}&{\bf 69.24}&{\bf 87.90} &1.8 days&8 G\\
				\bottomrule
			\end{tabular}
		\end{adjustbox}
	\end{center}
	\caption{Semantic segmentation results, training time (days) and GPU memory occupation (GBs) on Area 5 of the S3DIS dataset.}
	\label{tab:results_s3dis_area5}
\end{table}
\setlength{\belowcaptionskip}{-0.cm}

\subsection{Ablation study}
To better demonstrate the effectiveness of our proposed method, we conduct ablation studies on Area 5 of the S3DIS dataset to analyze the effects of the non-local operations on different levels. Tab.~\ref{tab:results_s3dis_area5} shows the segmentation accuracies with the non-local operation at different levels. 
The neighborhood-level non-local operation (PointNL$^1$) can obtain the mIoU of 60.03\% and is competitive with many point cloud segmentation approaches in recent years. 
When we further add the non-local operation at the superpoint level (PointNL$^{12}$), the mIoU can be improved by 2.15\%. 
When we cascade the non-local operations at three levels (PointNL$^{123}$), the mIoU can be further improved by 1.32\%. In addition, we also implement the original non-local operation, which directly applies the non-local operation on the whole points. In Tab.~\ref{tab:results_s3dis_area5}, the mIoU of the original non-local operation (baseline) is only 54.76\%, which is lower than that of PointNL. It implies that our hierarchical non-local model can effectively build semantic long-range dependencies between the points to obtain the discriminative local features of point clouds.

\subsection{Computational cost}
In terms of training time and GPU memory, we compare our method to PointWeb and the original non-local operation (baseline) on the Pytorch platform. For a fair comparison, on Area 5 in the S3DIS dataset, both codes are run on Tesla P40 GPUs for 100 epochs. The computational cost, GPU memory occupancy and segmentation accuracy of the compared methods are shown in Tab.~\ref{tab:results_s3dis_area5}.

For PointWeb~\cite{pc_zhao2019pointweb}, it needs at least 4.2 days, 24GB GPU memory and can obtain the mIoU of 60.28\%. The original non-local operation (baseline) costs about 6 days and 60G GPU memory, and obtain the mIoU of 54.76\%. While our method costs about 1.8 days, 8G GPU memory and can achieve mIoU of 63.50\%. 

The performance, computational cost and the GPU memory occupancy of the corresponding methods are shown in Tab.~\ref{tab:results_s3dis_area5}. The total number of parameters of our model is 3.5M. For testing, the inference time of each batch is 0.35s, while PointWeb requires 1.5s. Therefore, our method not only improves the performance of semantic segmentation but also greatly reduces the time cost and memory consumption.

\section{Conclusion}
We proposed a novel cascaded non-local neural network for point cloud semantic segmentation. In our proposed non-local neural network, we developed a new cascaded non-local module to capture the neighborhood-level, superpoint-level and global-level geometric structures of point clouds. By stacking the cascaded non-local modules, semantic context information of point clouds is propagated level by level so that the discriminativeness of local features of point clouds can be boosted. Experimental results on the benchmark point cloud segmentation datasets demonstrate the effectiveness of our proposed PointNL in terms of the segmentation accuracy and computational cost.

\addtolength{\textheight}{-12cm}   


{\small
	\bibliographystyle{IEEEtran}
	\bibliography{egbib}
}

\end{document}